\documentclass{article}
\usepackage[preprint]{log_2022}
\usepackage{booktabs}           % professional-quality tables
\usepackage{multirow}           % tabular cells spanning multiple rows
\usepackage{amsfonts}           % blackboard math symbols
\usepackage{graphicx}           % figures
\usepackage{duckuments}         % sample images

\usepackage[utf8]{inputenc} % allow utf-8 input
\usepackage[T1]{fontenc}    % use 8-bit T1 fonts
\usepackage{url}            % simple URL typesetting
\usepackage{booktabs}       % professional-quality tables
\usepackage{amsfonts}       % blackboard math symbols
\usepackage{nicefrac}       % compact symbols for 1/2, etc.
\usepackage{microtype}      % microtypography
\usepackage{xcolor}         % colors

\usepackage{soul}
\usepackage{ulem}

\usepackage{caption}
\usepackage{wrapfig}

\usepackage{footnote}
\makesavenoteenv{tabular}
\makesavenoteenv{table}

\usepackage{amssymb}
\usepackage{amsmath}
\usepackage{bm}
\newcommand{\vect}[1]{\boldsymbol{#1}}
\newcommand{\R}{\mathbb{R}}
\newcommand{\mat}[1]{\boldsymbol{#1}}

\usepackage{algorithm}
\usepackage[noend]{algorithmic}
\usepackage{pifont}
\usepackage{xcolor}
\newcommand{\xmark}{\color{red} \ding{55}}

\newcommand{\newrevised}[1]{{\color{black} #1}}
\newcommand{\oldrevised}[1]{{\color{black} #1}}
\newcommand{\revised}[1]{{\color{black} #1}}
\newcommand{\erase}[1]{}
\usepackage[numbers,compress,sort]{natbib}
\title{A Simple Way to Learn Metrics Between Attributed Graphs}

\author[Y. Kaloga et al.]{%
Yacouba Kaloga\\%\thanks{Equal contribution.}\\
\institute{Univ Lyon, Ens de Lyon, CNRS, \\ Laboratoire de Physique, UMR 5672\\ Lyon, France}\\
\email{yacouba.kaloga@ens-lyon.fr}\And
Pierre Borgnat\\%\thanks{Equal contribution.}\\
\institute{Univ Lyon, Ens de Lyon, CNRS, \\ Laboratoire de Physique, UMR 5672\\ Lyon, France}\\
\email{pierre.borgnat@ens-lyon.fr}\And
Amaury Habrard\\%\footnotemark[1]\\
\institute{University of Lyon, UJM-Saint-Etienne, \\CNRS, Laboratoire Hubert Curien, UMR 5516, \\ Saint-Etienne, France\\Institut Universitaire de France (IUF)}\\
\email{amaury.habrard@univ-st-etienne.fr}
}

\begin{document}

\maketitle

\begin{abstract}
The choice of good distances and similarity measures between objects
is important for many machine learning methods. Therefore, many
metric learning algorithms have been developed in recent years, mainly for Euclidean data, in order to improve performance of classification or clustering methods.
However, due to difficulties in establishing computable, efficient and
differentiable distances between attributed graphs, few metric learning algorithms adapted to graphs have been developed despite the strong interest of the community. 
In this paper, we address this issue by proposing a new Simple Graph Metric Learning - SGML - model with few trainable parameters based on Simple Graph Convolutional Neural Networks - SGCN - and elements of Optimal Transport theory. This model allows us to build an appropriate distance from  a database of labeled (attributed) graphs to improve the performance of simple classification algorithms such as $k$-NN. This distance can be quickly trained while maintaining
good performance as illustrated by the experimental studies presented
in this paper.
\end{abstract}

\section{Introduction}

Classification of attributed graphs has received much attention in recent years because graphs are well suited to represent a broad class of data in fields such as chemistry, biology, computer science, etc~\citep{shervashidze11,kersting16}. Advances were obtained, particularly thanks to the development of graph convolutional neural networks (GCN)~\citep{defferrard16,kipf16,xu18,wu19} of which many actually graph learning model can rely on~\citep{velickovic18,hamilton17i}. GCN has attracted interest in the recent years, due to their low computational cost, their ability to extract task-specific information, and their ease of training and integration into various models. Some tackle classification problems for attributed graphs by leveraging  GCN: they characterize and build Euclidean representations for attributed graphs both in a supervised (e.g.~\citep{xu18,ying18}) or unsupervised~(e.g.~\citep{sun20,bethune20}) way. Despite these achievements, classification methods based on direct evaluation of similarity measures between graphs remain relevant since they can obtain similar, and in some cases even better, performance~\citep{togninalli19}. Currently, most of these methods work in a task-agnostic way. However, given the diversity of graph datasets, we can not expect one similarity measure to be well suited for all of them, on all learning tasks. 

Adapting similarity measures to specific datasets and related tasks help to improve their generality and their performance. One of such approach is known as Metric Learning (hereafter ML), and has already been successful for Euclidean data. Xing et al. \cite{xing02} were the first to propose a Metric Learning method to improve a specific method and task ($k$-means for clustering of Euclidean data). This first work sparked a strong interest in ML which led to the development of many methods~\citep{goldberger05,weinberger09,bellet13,suarez18} for Euclidean data. In contrast, few of these methods exist for attributed graphs. Existing methods (e.g., \cite{yoshida21}) rely on iterative procedures which are hardly differentiable, and this makes also scalability an issue. 
In the state-of-the-art of classification, neural networks tend to currently dominate in the literature, yet building simple and learned (hence adapted to data and task) similarity measures between attributed graphs remain a relevant issue for at least two reasons: it allows to step up simpler graph classification algorithms, and also it allows to rely on graph kernels~\citep{kriege16,shervashidze11} which are, as of today, as efficient on numerous tasks as models relying on graph neural networks.  

\noindent \textbf{Our contribution.} To address the issue of scalability in Metric Learning for graphs, we propose here a novel graph ML method, called Simple Graph Metric Learning (SGML). In the first step, attributed graphs are coded as distributions by combining the attributes and the topology thanks to GCN. Then, relying on Optimal Transport, we define a novel \newrevised{\erase{similarity measure} metric} between these distributions, that we call Restricted Projected Wasserstein, $\mathcal{RPW}_2$ for short. $\mathcal{RPW}_2$ is differentiable and has a quasi-linear complexity on the distribution size (in number of bins; this is also the number of nodes); it removes certain limitations of the well known Sliced Wasserstein (noted $\mathcal{SW}_2$)~\citep{bonneel15}.\newrevised{\erase{The}} $\mathcal{RPW}_2$ \newrevised{\erase{similarity measure}} is then used to build a parametric \newrevised{\erase{distance} pseudo-metric} between attributed graphs which then has also a quasi-linear complexity on the graph size (in the number of nodes). The similarity measure proposed in SGML has a limited number of parameters, and it helps the model to scale efficiently. \oldrevised{Next, we focus on the the k-nearest neighbors (k-NN) method for classification.
An advantage of using k-NN is that, if the learning set grows, one can exploit it at near zero additional cost (since it only requires to store these new data) on the contrary of SVMs that would require to retrain the whole data (a task quadratic in size). Since many real datasets (e.g., graphs from social networks, or to detect anomalies on computer networks) are expected to have a growing size, this property is important for continual learning, and from an energetic and environmental stance to avoid costly retraining.
In order to use k-NN and train the distance, we propose a novel softmax-based loss function over class point clouds. It appears to be novel in the context of graph ML and it leads to  better results in the explored setting than the usual ML losses (i.e., those specifically built to improve k-NN for Euclidean data).
Our experiments show that SGML learns a metric increasing significantly the k-NN performance, compared to state-of-art algorithms for graph similarity measures.}

The article is organized as follows. In Section 2, we discuss related works on graph metric learning and on optimal transport theory applied to the construction of attributed graphs similarity measures.
Section~\ref{s:BK} provides useful notations and definitions  needed for the present work.
The SGML model is defined in Section~\ref{s:SGML_model}.
Finally, in Section~\ref{s:experiments}, we present various numerical experiments assessing the efficiency of our model.
These experiments show that in various conditions,  SGML  has great ability to build accurate distance with competitive performance with the state-of-the-art in classification of graphs, both in context of k-NN and kernel-based methods, and that despite its limited number of parameters.
A main advantage of the proposed SGML method is also its simplicity, hence leading to a scalable and efficient method for graph Metric Learning.
We conclude in Section~\ref{s:conclusion}. \\
\textbf{Societal Impact} The contribution is essentially fundamental, and we do not see any direct and immediate potential negative societal impact. \oldrevised{Conversely, the scalability of the method will help to alleviate the energy consumption of ML on graphs.}
\section{Related Works} 

\subsection{Graph Metric Learning} 
About ML for graphs, we can notably mention a series of works~\citep{bellet12,neuhaus07,jia21} that consist in learning a metric through Graph Edit Distance (GED). The major disadvantage of these methods is the complexity of the  computation of the  GED which can be only done for very small graphs.

Following the introduction of GCN, an approach based on Siamese neural networks has been proposed in~\cite{sofia18} for the study of brain connectivity signals, represented as graphs signals. In this specific case, all graphs are the same and they differ only by the signal they carry. This makes this method not applicable to most of datasets. More recently models without neural networks have been proposed: \cite{yoshida21} present \textit{Interpretable Graph Metric Learning} which builds a similarity measure by counting the most relevant subgraphs to perform a classification task. However, their method cannot handle large graphs. \cite{zhao19} proposes to learn a kernel based on graph persistent homology. The resulting model is also efficient, but it has the disadvantage of not being able to deal with discrete features in graphs.

As seen, existing work on graph ML are either limited by the assumptions made to build their model, or too costly, or not suitable to actually leverage simple (classification) algorithms and increase their performance. To obtain a simple graph ML procedure that is not itself too costly, we need to have a similarity measure between graphs that can be computed quickly. To construct such a distance, recent works suggest that Optimal Transport is an appropriate tool.

\subsection{Optimal Transport for Graphs} 

Optimal Transport (OT) has been put forward as a good approach to quickly compute  similarity measures between graphs, relying on the the fact that it provides tools for computing metric between distributions \citep{peyre20}. Recent studies have shown that efficient distances and kernels for graphs can be built from this theory. Fused-Gromov-Wasserstein~\citep{vayer19} is such a metric (distance in a mathematical sense) using OT to compare graphs through both their structures and attributes. Notably it allows one to compute barycenter of a set of graphs, and interpolation between graphs. Experimentally, it leads to good results in classification. \oldrevised{Its bi-quadratic complexity in the size of graphs is its main drawback, even if it can be reduced to cubic cost with entropic regularization.

In \cite{hermina19} an OT based approach to compare graphs is developed. It uses OT between specific signals on the graphs.} Thanks to a Gaussian distribution hypothesis, the analytical expression of the OT between these signals is derived. While the model provides good results, it is initially limited to graphs having the same size, and a task of node alignment (which has a cubic complexity) must be performed.
\cite{maretic22} relaxes the condition on size, yet the focus remains on graph alignment of non attributed graphs.

\cite{togninalli19} has proposed the Wasserstein Weisfeiler-Lehman (WWL) method which can be seen as an evolution of the previous work~\citep{hermina19} without the two hypotheses, neither on the size of the graphs nor on the specificity of the graph signals. In addition, a non trainable GCN is used to build task-agnostic characteristics which are then compared through OT. This pseudo-metric is then used to build an efficient kernel for graph classification. Unfortunately this model requires the computation of the optimal transport map which has a cubic cost (or quadratic with entropy regularization). 

While these models are efficient on classification tasks, their complexity remains high, and they are not fast enough (being quadratic or more) to be incorporated in a framework of Metric Learning. A part of our contribution is to provide such an optimal transport-based fast similarity measure for attributed graphs, with no restriction on the nature of the graphs (and their attributes) to be compared.

\section{Background on Metric Learning and Optimal Transport}
\label{s:BK}

\noindent \textbf{Notations.} Let us consider a finite dataset $\mathbb{X} = \{\vect{x}_i\}_{i=1}^{|\mathbb{X}|}$ 
whose elements are in $\R^q$. The dataset comes with a set of labels  $\mathbb{E} = \{e_i\}_{i=1}^{|\mathbb{E}|}$ and a labeling function $\mathcal{E}: \mathbb{X} \rightarrow \mathbb{E}$. 
We note $\mathcal{P}(\mathbb{X}) \subset \mathcal{P}(\mathbb{R}^{q})$ the set of discrete probability over $\mathbb{X} \subset \mathbb{R}^{q}$.
$\delta_{\vect{x}}$ is the Dirac distribution centered in $\vect{x}$. { We note $d$ a metric on $\mathbb{X}$. It verifies the following properties: \textbf{Symmetry} - $\forall (\vect{x},\vect{y}) \in \mathbb{X}^2, d(\vect{x},\vect{y}) = d(\vect{y},\vect{x})$; \textbf{Identity of indiscernibles} - $\forall (\vect{x},\vect{y}) \in \mathbb{X}^2, d(\vect{x},\vect{y}) = 0 \Leftrightarrow \vect{x} = \vect{y}$; \textbf{Triangle inequality} - $\forall (\vect{x},\vect{y},\vect{z}) \in \mathbb{X}^3, d(\vect{x},\vect{z}) \leq d(\vect{x},\vect{y}) + d(\vect{y},\vect{z})$. $d$ is referred to as a pseudo-metric when it follows these properties except the identity of indiscernibles. In this article, the term "distance" will be used sometimes in an informal way as a synonym of discrepancy or measures of similarity. \newrevised{Additionally the term "distribution" will always refer to discrete distribution.}}

\subsection{Learning a metric}
For ML, we suppose that a dataset $\mathbb{X}$ is given with the knowledge of two sets: $\mathcal{S}$ (similar) and $\mathcal{D}$ (dissimilar), containing pairs of some elements of $\mathbb{X}$. The goal is to build a parametric distance $d_\theta$ in such a way that the pairs of elements in $\mathcal{S}$ should be \textit{close} while the pairs in $\mathcal{D}$ should be \textit{far away}\footnote{Some algorithms use a third type of information, which consists of triples indicating that a given element must be closer to such  element than to  another element~\citep{bellet13}.}. These sets are often built from the labeling function of $\mathbb{X}$ such that $\{\vect{x}_i,\vect{x}_j\} \in \mathcal{S}$ if $\mathcal{E}(\vect{x}_i) = \mathcal{E}(\vect{x}_j)$ otherwise $\{\vect{x}_i,\vect{x}_j\} \in \mathcal{D}$. An optimization problem depending on $d_\theta$, $\mathcal{S}$ and $\mathcal{D}$ is then defined with a loss function $\mathcal{F}$ suitable for the purpose:

\begin{equation}
\label{eq:am_obj}
\max_{\theta} \mathcal{F}(d_\theta,\mathcal{S},\mathcal{D})
\end{equation}

We denote $\theta^{*}$ the optimal parameters.
The interest for building such a distance $d_{\theta^{*}}$ with respect to information in $\mathcal{D}$ and $\mathcal{S}$ lies in the fact that $\mathbb{X}$ is often included in a larger set, containing elements which are not labeled. The goal is that the obtained distance $d_{\theta^{*}}$ will ease learning algorithm to find these missing labels. 
A part of our work will be to introduce a new and suitable loss function $\mathcal{F}$ in metric learning literature for the problem of metric learning for graphs.

\subsection{Optimal transport}
Let us consider two finite datasets $\mathbb{X}$, $\mathbb{X}'$, and two distributions $\mu \in \mathcal{P}(\mathbb{X})$ et $\nu \in \mathcal{P}(\mathbb{X}')$ on these sets:
\begin{equation}
\label{eq:distribution_discrete}
\mu = \sum_{\vect{x}_i \in \mathbb{X}} a_i \delta_{\vect{x}_i}\text{\:\:\:and\:\:\:} 
\nu = \sum_{\vect{x}_i' \in \mathbb{X}'} b_i \delta_{\vect{x}_i'}
\end{equation}
 \noindent with $a_i \geq 0$, $b_i \geq 0$,
 $n=|\mathbb{X}|$, $n'=|\mathbb{X}'|$,
 and $\sum_{i=1}^{n} a_i = 1$, $\sum_{i=1}^{n'} b_i = 1$. Given a continuous cost function $c : \mathbb{R}^q \times \mathbb{R}^q \rightarrow \R_{+}$, one can build from optimal transport a metric between distributions with support in $\R^q$, the so-called 2-Wasserstein distance $\mathcal{W}_2$~:
\begin{equation}
\label{eq:wasse}
\begin{split}
\mathcal{W}_2(\mu,\nu) = 
\inf_{\pi_{i,j}\in \Pi_{a,b}}  {\Big(\sum_{i,j=1}^{n,n'} \pi_{i,j} c(\vect{x}_i,\vect{x}_j')^2}\Big)^{\frac{1}{2}}
\end{split}
\end{equation}
\noindent  $\Pi_{a,b}$ is the set of joint distributions on $\mathbb{X} \times \mathbb{X}'$, $\pi = \sum_{\substack{ i,j =1}}^{n,n'} \pi_{i,j} \delta_{(\vect{x}_i,\vect{x}_j')}$ whose marginals are the distributions $\mu = \sum_{\vect{x}_i'\in \mathbb{X}'} \pi(\cdot,\vect{x}_i')$ and $\nu = \sum_{\vect{x}_i\in \mathbb{X}} \pi(\vect{x}_i,\cdot)$. 
We note $\pi^{*} \in \Pi_{a,b}$ the optimal distribution (or coupling, or map) giving the solution of this problem. The cost function $c$ is taken as 2-norm: $c(\vect{x}_i,\vect{x}_j') = |\!|\vect{x}_i-\vect{x}_j'|\!|_2$, leading hence to the 2-Wasserstein distance. This defines an efficient way to compare distributions. One could use differentiable versions (w.r.t the parameters of a distribution) by considering the 1-Wasserstein~\citep{arjovsky17} or the entropic regularization of $\mathcal{W}_2$~\citep{cuturi13,peyre20}. Still, they are not suitable for metric learning because of the (initial) complexity (when $n = n'$) in $O(n^3 \log n)$, or $O(n^2 \log(n))$ with entropic regularization thanks to the Sinkhorn algorithm~\citep{peyre20}.

\noindent \textbf{Sliced Wasserstein distance ($\mathcal{SW}_2$).} In order to drastically reduce the cost for computing the OT, \cite{bonneel15} has proposed a modified metric $\mathcal{SW}_2$ which consists to compare the measures $\mu$ and $\nu$ via their one dimensional projections. Let $\vect{\theta} \in \mathbb{S}^{q-1}$ be a vector of the unit sphere of $\R^q$. Distributions $\mu$ and $\nu$ projected along $\vect{\theta}$ are denoted $\mu_{\vect{\theta}} = \sum_{\vect{x}_i \in \mathbb{X}} a_i \delta_{\vect{x}_i \cdot \vect{\theta}}$ and $\nu_{\vect{\theta}} = \sum_{x_i' \in \mathbb{X}'} b_i \delta_{\vect{x}_i \cdot \vect{\theta}}$. $\mathcal{SW}_2$ is defined as follows: 
\begin{equation}
\label{eq:swa}
\begin{split}
\mathcal{SW}_2(\mu,\nu)^2 = \int_{\mathbb{S}^{q-1}}\mathcal{W}_2(\mu_{\vect{\theta}},\nu_{\vect{\theta}})^2 d\vect{\theta} 
\end{split}
\end{equation}
The advantage of this formulation stems from the quasi-linearity in $n$ (or $n'$) of the computation cost of $\mathcal{W}_2$ distance between one dimensional distributions. The integral can be estimated via a Monte-Carlo sampling. The complexity is then (when $n' \leq n$) at most $O(M ( n \log n))$ with $M$ the number of samples (uniformly) drawn from $\mathbb{S}^{q-1}$. However, \cite{rowland19} shows that $\mathcal{SW}_2$ is a biased downwards compared to $\mathcal{W}_2$, since the  vector $\vect{\theta}$ for projection determines at the same time the OT plans and also the cost of transport; this leads to a less effective distance.

\noindent \textbf{Projected Wasserstein distance ($\mathcal{PW}_2$).} When $n=n'$, $\mathcal{PW}_2$ is introduced by \cite{rowland19} in answer to previous limitations. 
$\mathcal{PW}_2$ is computed similarly as $\mathcal{SW}_2$, but for each projection $\vect{\theta}$, the one dimensional optimal transport plan $\pi^{\vect{\theta},*}$ between $\mu_{\vect{\theta}}$ and $\nu_{\vect{\theta}}$ is used with the original distributions $\mu$ and $\nu$ so as to compute the transport cost:
\begin{equation}
\label{eq:pw}
\begin{split}
\mathcal{PW}_2(\mu,\nu)^2 = \int_{\mathbb{S}^{q-1}}  \sum_{i,j=1}^{n,n'} \pi_{i,j}^{\vect{\theta},*} |\!|\vect{x}_i-\vect{x}_j'|\!|_2^2 d\vect{\theta} 
\end{split}
\end{equation}

They show that this formulation \newrevised{\erase{gives} defines} a metric, has good properties and is more suitable for several learning tasks, e.g. generative tasks or reinforcement learning. Unfortunately their result holds only for uniform distributions of the same size. \newrevised{Our method rely on an extended version of this definition, involving distributions of different sizes and not necessarily uniform. \erase{We extend the method to distributions of different sizes.}}

\section{A Simple and Scalable Graph Metric Learning}
\label{s:SGML_model}
Let us consider a dataset $\mathbb{G}$ of attributed graphs with labeling set $\mathbb{E}$ and labeling function $\mathcal{E}$. For a given graph $\mathcal{G} \in \mathbb{G}$ having $\vect{A}$ as adjacency matrix, we call $n$ the number of node of the graph. Each node $i$ of $\mathcal{G}$ carry features $\vect{X}(i,:) \in \R^q$; thus $\vect{X} \in \R^{ n \times q }$ is the matrix of attributes of the graph.

\subsection{From graph to distribution}
Previous works using OT (pseudo-)metric have shown that comparing graphs through the signal they carry is a good way to compare them; we follow this path. The first step of our learning method consists in the generation of features jointly representative of the structure of each graph $\mathcal{G}$ and the attributes of their nodes $\vect{X}$. We use for this purpose Simple GCN~\citep{wu19}, a streamlined version of GCN in which all the intermediate non-linearities have been removed. This choice is dictated by the need to strongly reduce the number of trainable parameters, and it accelerates the training without degrading its performance compared to other GCN. This Simple GCN creates features as: 
\begin{equation}
\vect{Y} =  \text{ReLU}( \widetilde{\vect{A}}^r \vect{X}  \vect{\Theta}  )
\label{eq:sgcn_it2}
\end{equation}
where $\vect{X} \in \R^{ n \times q }$ are the initial attributes of the nodes, $\widetilde{\vect{A}} = \vect{A} + \vect{I}_n$ (where $\vect{I}_n$ is the identity matrix of $\mathbb{R}^n$) and $\vect{Y} \in \R^{ n \times p }$ are the features computed by SGCN. The neighborhood exploration depth $r$ of this GCN is one of the hyperparameters of the method, along with the dimension $p$ of the extracted features  $\vect{Y}$. The coefficients of the matrix $\vect{\Theta} \in \R^{q \times p}$ of this GCN are the (only) trainable weights of the method. We will always choose $p \leq q$, so the method has at most $q^2$ trainable parameters. From the extracted features $\vect{Y}$, we define a uniform distribution whose suport is the nodes' characteristics:
\begin{equation}
\mathcal{D}_{\vect{\Theta}}(\mathcal{G},\mathbf{X}) = \sum_{i = 1}^n \frac{1}{n} \delta_{\vect{Y}(i,:)}
\label{eq:sgcn_dist}
\end{equation}

\noindent This first step is similar to WWL \citep{togninalli19}, {except}  that we consider a trainable GCN, $\vect{\Theta}$ being the trainable parameters. 
In eq.~(\ref{eq:sgcn_dist}), both the structure $\mathcal{G}$ and the attributes $\mathbf{X}$ are accounted for.
Next, we propose a novel way to evaluate the similarity between attributed graphs using these distributions.

\subsection{From distributions to distance}

The distances between graphs are computed as a distance between their representative distributions (Eq.~(\ref{eq:sgcn_dist})) with OT; specifically, we propose a novel one, called Restricted Projected Wasserstein (and noted $\mathcal{RPW}_2$) extending $\mathcal{PW}_2$ previously introduced in \cite{rowland19}.

\noindent \textbf{Restricted Projected Sliced-Wasserstein.} In \cite{rowland19}, $\mathcal{PW}_2$ is only defined for uniform distributions when $n=n'$. We extend this \newrevised{definition} to cases $n \neq n'$, \newrevised{and we remove the constraint of uniformity. }
\newrevised{We show in Appendix~\ref{annexe_rpw} that it still defines a metric (on discrete distribution space). \erase{While the symmetry and the identity of indiscernibles is still verified, there is no guarantee that $\mathcal{PW}_2$ remains a metric on uniform distribution space, because the triangle inequality cannot be derived as easily as when $n = n'$.}}

% \revised{\st{}

In order to compute this quantity, we could rely on Monte-Carlo sampling, and the complexity would be $O(M p n \log(n))$. This can be prohibitive due to the term $pM$. In order \newrevised{to have an even more scalable \erase{obtain a scalable}} model, we restrict the projections to be alongside the basis vectors $\{\vect{u}_k\}_{k=1}^{p}$ of $\mathbb{R}^{p}$ only. \revised{This choice stems from a spanning constraint that allows us to define a quantity \newrevised{(named Restricted $\mathcal{PW}_2$, or $\mathcal{RPW}_2$ for short)} verifying the identity of indiscernibles without increasing significantly the computing time (see Appendix~\ref{annexe_rpw}). \newrevised{This \erase{property} guarantees that $\mathcal{RPW}_2$  is also a metric of discrete distribution space as $\mathcal{PW}_2$ (see also Appendix~~\ref{annexe_rpw})}. Therefore it can always distinguish distributions that are different but also (associated with the continuity) that, when two distributions are getting closer, then $\mathcal{RPW}_2$ tends towards 0; this is important in ML context.} %\st{Our experiments will show that this choice was relevant.} 
\newrevised{$\mathcal{RPW}_2$ is expressed as: \erase{This choice defines a new distance, called Restricted $\mathcal{PW}_2$, or $\mathcal{RPW}_2$ for short, reading as:}}
\begin{equation}
\label{eq:rpw}
\begin{split}
\mathcal{RPW}_2(\mu,\nu)^2 =  \frac{1}{p} \sum_{k = 1}^{p}  \sum_{i,j=1}^{n,n'} \pi_{i,j}^{ {\vect{u}_k},*} |\!|\vect{x}_i-\vect{x}_j'|\!|_2^2  
\end{split}
\end{equation}
\newrevised{Note that $\mathcal{RPW}_2$ defines a metric therefore it should considered as such and not as an attempt to approximate $\mathcal{PW}_2$.} \newrevised{A major \erase{The main}} advantage of $\mathcal{RPW}_2$ is that it is defined by a deterministic formula; this avoids the variability introduced by a Monte-Carlo sampling \newrevised{(when one would need to evaluate $\mathcal{PW}_2$ or $\mathcal{SW}_2$)}. 
\revised{ \newrevised{Anyway,} we can notice that for a given $\vect{u}_k$, many $\pi_{i,j}^{{\vect{u}_k}}$ may be optimal for the projected distribution on $\vect{u}_k$, while they may lead to different values when computing Eq.~(\ref{eq:rpw}).
In order to have an unambiguous and deterministic definition, in such cases we can choose among admissible optimal transport maps the one which minimizes Eq.~(\ref{eq:rpw}). \newrevised{In fact it is even necessary for it to be a metric.} However since this case is quite rare, in our implementation we simply took the first one returned by our sorting algorithm.} The complexity of $\mathcal{RPW}_2$ is given by $O( p^2 n \log(n) )$ which saves a factor $\frac{M}{p}$ as compared to  $\mathcal{PW}_2$ and this term is often greater than 10. \newrevised{\erase{Finally, let us note that we did not find numerically evidences that $\mathcal{RPW}_2$ does not verify the triangle inequality; we only found a few examples of triplets where, numerically, the inequality was not satisfied at the level of the numerical precision limit. However further work is needed to answer this question.}}

\newrevised{From $\mathcal{RPW}_2$}, we define a parametric distance $d_{\vect{\Theta}}^{\mathcal{RPW}_2}$ between two attributed graphs $(\mathcal{G},\mathbf{X})$ and $(\mathcal{G}',\mathbf{X}')$~: 
\begin{equation}
\label{eq:grpw}
\begin{split}
d^{\mathcal{RPW}_2}_{\vect{\Theta}}(\mathcal{G},\mathcal{G}') = \mathcal{RPW}_2(\mathcal{D}_{\vect{\Theta}}(\mathcal{G},\mathbf{X}),\mathcal{D}_{\vect{\Theta}}(\mathcal{G}',\mathbf{X}'))
\end{split}
\end{equation}

All the metric learning experiments will be conducted using this distance, excepted in an ablative study where we report the use of $\mathcal{SW}_2$ and $\mathcal{PW}_2$. \newrevised{Note that $d_{\vect{\Theta}}^{\mathcal{RPW}_2}$ is a pseudo-metric, since there are different graphs that can have the same features outputted by the GCN, hence leading to similar representative distribution, therefore the identity of indiscernibles is not verified.}

\subsection{Loss for training distance: the Nearest Class Cloud Metric Learning}        
\label{ss:loss}
The last element to complete our model is to define the loss function $\mathcal{F}$ for Eq.~(\ref{eq:am_obj}). We propose here a new loss function for the purpose of improving the $k$-nearest neighbors method. Actually there are classical losses already efficient for this purpose: one can notably mention Large Margin Nearest Neighbor (LMNN) \citep{weinberger09} and Neighbourhood Component Analysis (NCA)\citep{goldberger05}. 
However, the optimization is done using a gradient descent algorithm.
Since computing all pairwise distances between graphs at each step of gradient descent would be intractable for large datasets, we have to train our loss in a batch way. In this context, LMNN may be not relevant since this method works locally and a batch is often not representative of the true neighborhood of an element of the dataset. 
On the contrary NCA loss can be trained in a batch way, as it relies on a probability model which tends to attract elements with the same label with each other, wherever they are.
\oldrevised{However, preliminary experiments showed only a slight improvement of the k-NN with NCA. Therefore we have constructed a new loss which proposes a different way to ensure the same condition (see Appendix~\ref{ann:loss}) and which experimentally works better in our setting (see Ablative study, Sec.~\ref{sec:abla}).} The model is called Nearest Cloud Class Metric Learning  (NCCML); the probability of being labeled by $e \in \mathbb{E}$ for a graph $\mathcal{G}$ depends on the distance to the point clouds of a class (hence the name of the method):
\begin{equation}
\label{eq:ncmml}
p_{\mat{\Theta}}(e | \mathcal{G}) = \frac{\exp \Big({ - \sum_{\substack{\mathcal{G}_i \in \mathbb{G}\\ \mathcal{E}(\mathcal{G}_i)  = e}}d_{\mat{\Theta}}^{\mathcal{RPW}_2} {(\mathcal{G},\mathcal{G}_i)^2}}\Big)}{ \sum_{e' \in \mathbb{E}} \exp \Big({ - \sum_{\substack{\mathcal{G}_i \in \mathbb{G}\\ \mathcal{E}(\mathcal{G}_i ) = e'}}d_{\mat{\Theta}}^{\mathcal{RPW}_2} {(\mathcal{G},\mathcal{G}_i)^2}}\Big)}.
\end{equation}

 Given this probability, we want to construct the distance $d_{\vect{\Theta}}^{\mathcal{RPW}_2}$ maximizing the probability that the labeled graphs in the dataset have the correct labels, which leads to solve the following problem:

\begin{equation}
\label{eq:ncmml2}
\max_{\mat{\Theta}} \mathcal{F}^{\mathbb{G}}_{\Theta} = \max_{\mat{\Theta}}  \sum_{\mathcal{G}_i \in \mathbb{G} , \mathcal{E}(\mathcal{G}_i) \neq \emptyset} \log p_{\mat{\Theta}}( \mathcal{E}(\mathcal{G}_i) | \mathcal{G}_i).
\end{equation}

\oldrevised{By maximizing this loss, we construct a distance which, for each element, favors its relative distance to elements of the same labels compared to those of different labels. This should favor k-NN, especially when $k > 1$.}  We will show in the experiments that, in this specific context, NCCML exhibits better performance than NCA. 
\oldrevised{More details on NCCML can be found in Appendix~\ref{ann:loss}}.
  
\begin{algorithm}[t]
\caption{SGML: High-level algorithm to build $d^{\mathcal{RPW}_2}_{\vect{\Theta}^{*}}$.}  
\label{algo:nccml}
\begin{algorithmic} 

\REQUIRE A dataset of attributed graphs $\mathbb{G}$ and their labeling function $\mathcal{E}$.

\FOR{each epoch $e \in \{1,\dots,E\}$}
\STATE Build a partition: $\cup_k B_k = \mathbb{G}$ such that $B_k \cap B_{k'} = \emptyset$.
\FOR{each batch $B_k$}
\FOR{each graph pair $(\mathcal{G},\mathcal{G}')$ $\in\:$ $B_k \times B_k$}
\STATE Compute distance $d^{\mathcal{RPW}_2}_{\Theta}(\mathcal{G},\mathcal{G}')$ (Eq.~(\ref{eq:grpw}))
\ENDFOR
\STATE Compute $- \mathcal{F}_{\vect{\Theta}}^{B_k}$ (Eq.~(\ref{eq:ncmml2})) and apply an iteration of Adam descent algorithm.
\ENDFOR
\ENDFOR
\RETURN all pairwise distance $d^{\mathcal{RPW}_2}_{\vect{\Theta}^{*}}$ in $\mathbb{G}$.
\end{algorithmic}
\end{algorithm}

\subsection{ Computational aspects}

\label{sec:ca}

 We test, in the next Section, the proposed metric learning method with $\mathcal{RPW}_2$ (and $\mathcal{SW}_2$ or $\mathcal{PW}_2$ in ablative studies).

\textbf{Optimization.} In terms of optimization, we can differentiate directly with respect to one dimensional distribution parameters of Wasserstein distance, thus we can also differentiate through $\mathcal{RPW}_2$ (Eq.~(\ref{eq:rpw})) (and also \newrevised{approximation} of $\mathcal{SW}_2$ (Eq.~(\ref{eq:swa})) or $\mathcal{PW}_2$). Self-differentiation techniques can be used on these expressions~(see~\citep{peyre20}). We implemented our algorithm in \texttt{tensorflow}\footnote{The implementation can be found in the supplementary material.}. The minimization of the loss is performed by \textit{batch} and stochastic gradient descent (in particular with the optimizer \textit{Adam}~\citep{kingma14}).

\textbf{Parameters.} 
The following default parameters are used (unless otherwise indicated in the text): learning rate $l_r = 0.999*10^{-2}$, number of epochs $E = 10$, batch size $B = 8$, and the GCN output features size $p = \min(5,q)$. For experiments involving $\mathcal{SW}_2$ and $\mathcal{PW}_2$, the sampling number is set to $M=50$ which is a common value used in the literature.
 
\noindent \textbf{Time complexity.} Theoretically, the training time is negligible compared to the computation of all pairwise distances; therefore we focus on this last step for the time complexity analysis (see Appendix~\ref{annexe_runtimes} for runtimes per dataset). If we denote $\tilde{n}$ the number of average  nodes of a graph, the total complexity of this computation with $\mathcal{RPW}_2$ (resp. $\mathcal{SW}_2$) is given by $O( { |\mathbb{G}| \tilde{n}( p^2 +\tilde{n} r p) }+  |\mathbb{G}|^2 p^2 \tilde{n} \log \tilde{n})$  (resp. $O({|\mathbb{G}| \tilde{n}( p^2 +\tilde{n} r p) }+  |\mathbb{G}|^2 pM \tilde{n} \log \tilde{n})$). The first terms occur for application of GCN and the latest for computing distances. In practice, for not too large $\tilde{n}$ values, a quadratic implementation exploiting vectorization can be faster (see section~\ref{sec:rtc}). \oldrevised{Furthermore, one can see that the GCN becomes the limiting element for scaling (on graph sizes); in practice, the sparsity of the adjacency matrix and the optimizations on GPUs limit this problem. However, it is still an active research topic to determine the less expensive ways to characterize the nodes~\cite{Bojchevski20,frasca20}}. 

\textbf{Spatial Complexity.} Our quadratic implementation mentioned above requires to store in memory a tensor of size $O(\tilde{n}^2p)$ for $\mathcal{RPW}_2$ and $O(\tilde{n}^2M)$  for $\mathcal{SW}_2$ or $\mathcal{PW}_2$. The sequential implementation has a $O(\tilde{n})$ spatial complexity (more details on these implementations are in Appendix~\ref{ann:imp}). Anyway for both implementations, for the dataset of graphs considered, SGML is very cheap in terms of memory consumption in regards of actual GPU capability.

\section{Experiments}
\label{s:experiments}

\subsection{Datasets}

For the experiments, we use a large panel of data sets from the literature~\citep{kersting16}\footnote{\url{http://graphkernels.cs.tu-dortmund.de}}: ENZYMES, PROTEINS, IMDB-B, IMDB-M, MUTAG, BZER, COX2 and NCI1. More information on these datasets can be found in Appendix~\ref{app:datasets}. Additional details about the following experiments can be found in Appendix~\ref{ann:adddetails} for reproducibility.  \oldrevised{When a dataset has discrete features, they are one-hot encoded.}

\subsection{$\mathcal{RPW}_2$ Running times} 

\begin{figure}[h]
    \centering
    \includegraphics[width=10.5cm,height=29cm,keepaspectratio]{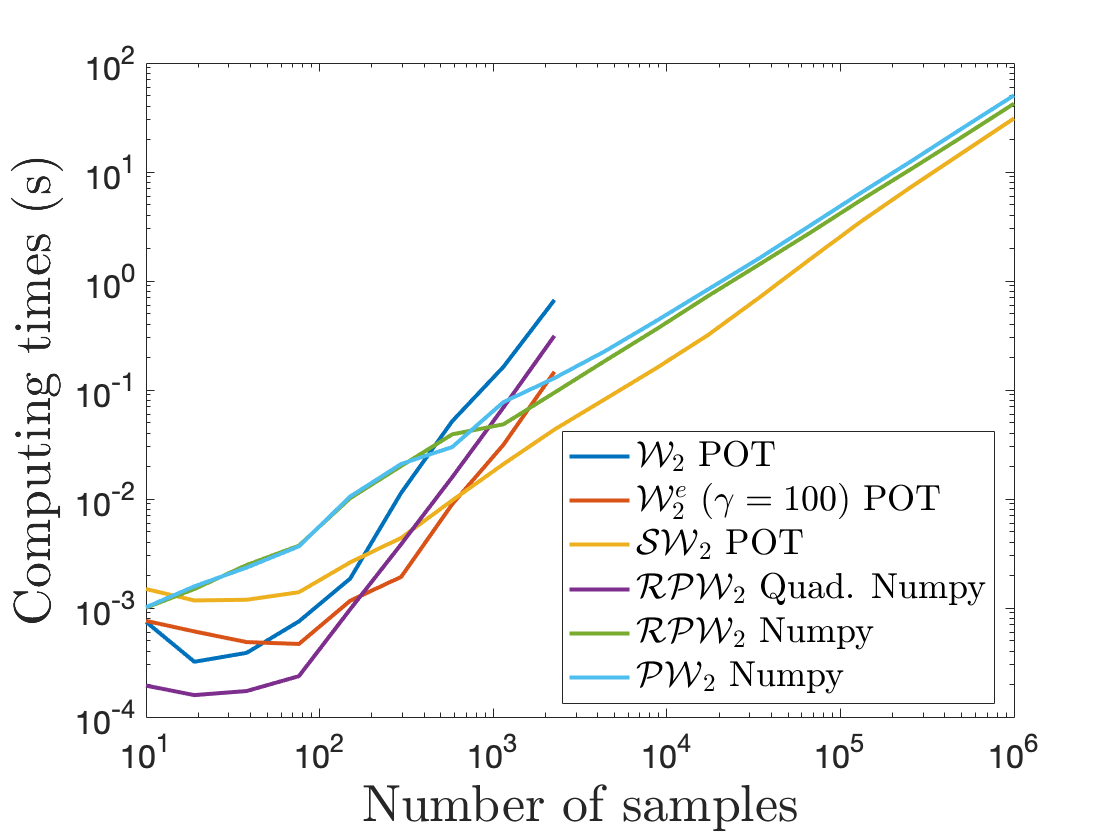}
    \caption{\color{black} \textbf{Run time comparisons.}}
    \label{fig:rnt}
\end{figure}

\label{sec:rtc} 

We have generated uniform random (normal) distributions with support in $\mathbb{R}^5$ of size ranging from $10^1$ to $10^6$. \oldrevised{These sizes of the distributions correspond to graph sizes $n$ (number of nodes). The choice of $\mathbb{R}^5$ is motivated by the usual good performance of ML when performed in small dimension.} We compare the running time to compute the distance between these distributions with $\mathcal{W}_2$, $\mathcal{W}_2^e$, ($\mathcal{W}_2$ with entropic regularization parameter $\gamma = 100$), $\mathcal{SW}_2$ using POT~\cite{flamary2021pot} library, \newrevised{$\mathcal{PW}_2$} and $\mathcal{RPW}_2$. For $\mathcal{RPW}_2$ we compare both the quadratic and the sequential (\texttt{numpy}) implementations we developed. \oldrevised{The results can be found on Figure~\ref{fig:rnt}. Additional details and results are given in Appendix~\ref{annexe_runtimes2}.}

As expected $\mathcal{SW}_2$, \newrevised{$\mathcal{PW}_2$} and $\mathcal{RPW}_2$ are the methods scaling the best: we obtain the expected (quasi) linear slope for \newrevised{all three} methods $O(n \log n)$. As soon as $n>10^{4}$, \newrevised{These three methods \erase{$\mathcal{SW}_2$ and $\mathcal{RPW}_2$} allows \erase{us}} to compute distances between distributions of several orders of magnitude larger for the same time as $\mathcal{W}_2$ and $\mathcal{W}_2^e$. Although $\mathcal{SW}_2$, \newrevised{$\mathcal{PW}_2$} and $\mathcal{RPW}_2$ scale mostly the same, $\mathcal{SW}_2$ seems a bit faster than $\mathcal{PW}_2$ and $\mathcal{RPW}_2$. \newrevised{However, the slope of $\mathcal{RPW}_2$ is a little bit smaller than $\mathcal{SW}_2$ one. (See Appendix~\ref{annexe_runtimes2}.) Anyway, \erase{However}} we will show in the next experiment (Sec.~\ref{sec:abla}) that $\mathcal{RPW}_2$ builds better metrics than $\mathcal{SW}_2$. Finally, \newrevised{\erase{we can}} note that the quadratic implementation \newrevised{of $\mathcal{RPW}_2$} is the fastest for samples with less than 200 instances, which is the case for the datasets considered in the following experiments.

\begin{table}[h!]
  \caption{{\textbf{Results of the main experiments for datasets of graphs with discrete attributes.} Features are node labels for NCI1, PROTEINS and ENZYMES; and degrees for others. Accuracy is in bold green when it is the best of its block. For $\mathcal{FGW}$-WL (resp. PSCN), depth is set to 4 (resp. 10).}} 
\label{data_results}

 \setlength{\tabcolsep}{2.25pt}
 \fontsize{9}{10}\selectfont
 
  \centering
  \begin{tabular}{lcccccc}
    \toprule
 
Method     & MUTAG    & NCI1   & PROTEINS & ENZYMES   & IMDB-M &  IMDB-B \\  
 \midrule
 \multicolumn{7}{c}{\textbf{k-NN}} \\
  \midrule
$\bm{\mathcal{RPW}_2}$ &   \color{green}  $\bm{90.00\pm 7.60}$     &  $72.12\pm1.65$& \color{green} $\bm{70.18\pm4.01}    $    & \color{green} $49.00\pm8.17$ & \color{green}  $45.00\pm5.46$ &$68.90\pm5.45$ \\  
Net-LSD-h  &   84.90   & 65.89 & 64.89  & 31.99& 40.51&$68.04$\\
FGSD    &  86.47   &  \color{green}75.77   & 65.30 & 41.58 & 41.14 &\color{green} 69.54 \\
NetSimile      &   84.09      &  66.56   & 62.45    & 33.23  & 40.97& 69.20\\ 
 \midrule
  \multicolumn{7}{c}{\textbf{SVM $\&$ GCN }} \\
  \midrule
 $\bm{\mathcal{RPW}_2}$   &   \color{green}  $\bm{88.95\pm7.61}$      &   $74.84\pm1.81$     & $74.55\pm 4.19$   & $54.00\pm7.07$ &   \color{green}  $51.00\pm5.44$ & \color{green}  $72.00\pm3.16$\\  
WWL   &           $87.27\pm1.50$ &     $85.75\pm0.25$    &   $74.28\pm0.56$  &  \color{green} $\bm{59.13\pm0.80}$  & \xmark & \xmark \\ 
$\mathcal{FGW}$     &   $83.26\pm10.30$     &   $72.82\pm1.46$    &    \xmark    & \xmark  & $48.00\pm3.22$ &  $ 63.80\pm3.49 $  \\ 
$\mathcal{FGW}$-WL &  $88.42\pm5.67$     &   \color{green}   \color{green} $\bm{86.42\pm1.63}$    &   \xmark    & \xmark & \xmark & \xmark \\   
WL-OA &   $87.15\pm1.82$      &  $86.08\pm0.27$       &  \color{green} $\bm{76.37\pm0.30}$  & $58.97\pm0.82$   & \xmark & \xmark \\  
PSCN      &   $83.47\pm10.26$     &$70.65\pm2.58$  &  $58.34\pm7.71$ & \xmark   & \xmark& \xmark \\ 
    \bottomrule
  \end{tabular}
\end{table}

\begin{table}[h!]
  \caption{\textbf{Results of the main experiments for datasets of graphs with continuous attributes graphs datasets.} The best accuracy are in bold green. Note that for PROTEINS, ENZYMES and CUNEIFORM we concatenate continuous attributes with discrete attributes to build an extended continuous attributes (see Appendix \ref{ann:adddetails} for more details).     }
 \label{data_results_con}
  \centering
   \setlength{\tabcolsep}{2.25pt}
 \fontsize{9}{10}\selectfont
  \begin{tabular}{llllll}
    \toprule

Method & BZR   & COX2  & PROTEINS   & ENZYMES  & CUNEIFORM

\\ 
\midrule

$\bm{\mathcal{RPW}_2}$ (kNN)   &    \color{green} $\bm{85.61\pm2.98}$     & \color{green}  $\bm{79.79\pm2.18}$ &  $71.79\pm4.47$  &  $51.66\pm5.16$ &  $54.81\pm12.26$   \\ 
 
      \midrule
  \multicolumn{6}{c}{\textbf{SVM $\&$ GCN }} \\
  \midrule
          
$\bm{\mathcal{RPW}_2}$    &   $84.39\pm3.81$    &    \color{green}  $\bm{78.51\pm0.01}$ & $74.29\pm4.11$   &  $48.83 \pm 4.78$ &  $64.44\pm10.50$      \\ 

WWL   &    $84.42\pm2.03$    &   $78.29\pm0.47$ & \color{green} $\bm{77.91\pm0.80}$ & \color{green} $73.25\pm0.87$ &  \xmark   \\ 
$\mathcal{FGW}$  &  \color{green}  $\bm{85.12\pm4.15}$      &   $77.23\pm4.86$ & $74.55\pm2.74$&  $71.00\pm6.76$ &\color{green}   $76.67\pm7.04$   \\ 

PROPAK    &   $79.51\pm5.02$   &  $77.66\pm3.95$   & $61.34\pm4.38$& $71.67\pm5.63$ & $12.59\pm6.67$     \\ 
HGK-SP    &   $76.42\pm0.72$    &   $72.57\pm1.18$    & $75.78\pm0.17$ & $66.36\pm0.37$ &  \xmark    \\ 
 
PSCN [K = 10] (GCN)  & $80.00\pm4.47$        &   $71.70\pm3.57$& $67.95\pm11.28$  & $26.67\pm4.77$ &  $25.19\pm7.73$  \\ 
\bottomrule
  \end{tabular}
\end{table}

\subsection{Supervised classification}
We evaluate the method in two ways: by using k-NN directly on the computed distances, and by using a SVM with a custom kernel built from the model proposed. We eventually compare the method to several (pseudo-) metric and distances from literature such as NetLSD~\citep{Tsitsulin18}, WWL~\citep{togninalli19},
 $\mathcal{FGW}$~\citep{vayer19}.

\noindent \textbf{k-Nearest Neighbors.} Datasets are split in a training (90\%) and test set (10\%). For each of them we train $\mathcal{RPW}_2$ following Algorithm~\ref{algo:nccml} (in the appendix) on the training set with only one hyperparameter to adjust: the depth of SGCN taken as $r=\{1,2,3,4\}$ for all datasets, except for MUTAG for which we go up to $7$. 
The training is done for each  parameter $r$ during 10 epochs. A 5-fold cross validation of the number of neighbors $k = \{1, 2, 3, 5 ,7\}$ to be considered is performed on the training set using the considered distance. Then for the best $k^{*}$, we keep the associated validation accuracy, and we finally train a k-NN on the whole training set and evaluate its accuracy on test set. This experiment is averaged on 10 runs. The final test accuracy retained is the one associated to the largest validation accuracy. In this procedure, test set labels were never seen during neither training nor validation. Results are given in the first lines of Table~\ref{data_results} for graphs having labeled nodes and of Table~\ref{data_results_con} for graphs with continuous attributes.

The learning metric framework combined with k-NN allows us to obtain good performance in classification tasks, in particular for datasets of graphs with continuous attributes. \oldrevised{The exception is ENZYMES where we can see a lower net performance.} For discrete attributes, SGML performs slightly below the state-of-the-art, yet it outperforms the existing distances classically combined with k-NN.  Experiments show that our graph ML distance framework is efficient.

\noindent \textit{\textbf{Note:} This procedure is very similar to the one used by WWL, except that the parameter $k$ is replaced by the corresponding parameters of their kernel (see next section).}

\noindent \textbf{SVM.} To compare to graph kernel methods, the experiment described in the previous section is reproduced using a SVM for classification. The kernel $\vect{K}_{\mathcal{RPW}_2} = \exp(- \lambda d_{\Theta^{*}}^{\mathcal{RPW}_2})$ is built from the constructed distance. In this experiment, kernel hyperparameter $\lambda$ and SVM hyperparameter $C$ are tuned similarly as the parameter $k$ above. The set of possible $\lambda$ (resp. $C$) values are 6 (resp. 12) regularly spaced values between $10^{-4}$ and $10^{1}$ (resp. $10^{-4}$ and $10^{5}$ including 1). The results are provided in Table~\ref{data_results} (bottom part).

In this part of the table, one can see that the distance learned with our model performs as well as other OT distances when used as a kernel, on the majority of the datasets. We reach or are slightly above state of the art results on 5 datasets over 6 but are still below on NCI1. We recall that our method is specifically designed for the k-nearest neighbors method and that its computational complexity  is much lower than  many of the best methods on these datasets (notably WWL and $\mathcal{FGW}$).

\subsection{Ablative study}
\label{sec:abla}
We perform experiments to justify the design choice of our model. Specifically we show that these choices effectively help to improve k-NN performance by reproducing the experiments above (with k-NN) on different versions of the method without some (or all) of our propositions.

\begin{table}[h!]
  \caption{\textbf{Ablative study results.} Acc. is the accuracy. $\Delta$ is the difference in accuracy between the model of the column and the proposed one SGML whose results are on Table.~\ref{data_results}. Red negative (resp. Green positive) number means that our model performs better (resp. worse). }
\label{data_results2}

 \setlength{\tabcolsep}{2.25pt}
 \fontsize{9}{10}\selectfont

  \centering
  \begin{tabular}{lllllllll}
    \toprule
  Dataset     & \multicolumn{2}{c}{\textbf{WWL}} & \multicolumn{2}{c}{\textbf{SGML} - $\bm{\mathcal{SW}_2}$} & \multicolumn{2}{c}{\textbf{SGML} - NCA}& \multicolumn{2}{c}{\textbf{SGML} - $\bm{\mathcal{PW}_2}$}  \\
  \cmidrule(r){2-3} 
  \cmidrule(r){4-5}
  \cmidrule(r){6-7}
  \cmidrule(r){8-9}
  Method & \textbf{Acc.}   & $\bm{\Delta}$ & \textbf{Acc.}   & $\bm{\Delta}$   & \textbf{Acc.}   & $\bm{\Delta}$& \textbf{Acc.}   & $\bm{\Delta}$ \\
  \midrule
  \textbf{BZR}    & 78.05     &  \color{red} - 7.56   & $82.93$   &  \color{red}  - 2.68     &  83.41   & \color{red} - 2.20   & 84.39 & \color{red}  - 1.22 \\ 
\textbf{COX2}   &  78.51    &   \color{red}  -1.26  & 78.30    &  \color{red}  - 1.49  &  77.66  & \color{red} - 2.13  & 78.94 &\color{red} -  0.85 \\
\textbf{MUTAG}  &   83.68   & \color{red}  - 6.32     &  86.84    &  \color{red}  - 3.16     &  87.37& \color{red} - 2.63  & 90.00 & \color{orange} 0.00 \\ 
\textbf{NCI1} &    80.43   & \color{green}  5.31   &  69.03    &  \color{red} - 3.09 &   69.66  & \color{red}   - 2.46  & 72.90 & \color{green} 0.78\\ 
\textbf{PROTEINS}  &   71.60     &    \color{green} 1.42  &71.34    & \color{green} 1.16  &  71.70 &  \color{green}1.52 & 70.54 & \color{green} 0.36  \\ 
\textbf{IMDB-B}  &   68.20     &  \color{red} - 0.7  & 68.20    & \color{red} -0.70  &  67.40 & \color{red} -1.5 & 68.80 &\color{red}   - 0.10 \\ 
\textbf{IMDB-M}  &   48.73     &   \color{green}3.73  & 42.33    & \color{red}-2.67  &  42.73 &  \color{red}-2.27 & 44.13 & \color{red}   - 0.87 \\ 
\textbf{ENZYMES} &  56.00 &  \color{green} 7.00  &  44.33    &  \color{red} - 4.67    &   55.33    &   \color{green}6.33 &  44.83 & \color{red} -4.17 \\ 
    \bottomrule 
  \end{tabular}
\end{table}

\noindent \textbf{Raw model.} Without any of our novel propositions,
the method would be equivalent to WWL, which corresponds to use the Wasserstein distance between distributions of Eq.~(\ref{eq:sgcn_dist}), where $\vect{Y}$ is generated with GIN~\citep{xu18}, a non trainable GCN. This specific case corresponds to the first column denoted \textbf{WWL} of Table~\ref{data_results2}.
We see that even if there are datasets where there is a loss of performance, others benefit from the learned metrics. \oldrevised{Moreover we remind that our distance is much less expensive to use than $\mathcal{W}_2$ on which WWL is based.}

\noindent \textbf{SGML with $\bm{\mathcal{SW}_2}$.} This second ablative study is in the second column, denoted \textbf{SGML}-$\bm{\mathcal{SW}_2}$, of Table~\ref{data_results2}, and is related to replacing $\mathcal{RPW}_2$ by $\mathcal{SW}_2$.
The result clearly validates our choice to use  $\mathcal{RPW}_2$ instead of $\mathcal{SW}_2$. Our model is the best one except on one  dataset.

\noindent \textbf{SGML with $\bm{NCA}$.} For this experiment we replaced the loss NCCML by the NCA loss. The result is in the third column, \textbf{SGML} - NCA of Table~\ref{data_results2}.
It appears that NCCML is often more appropriate than NCA in our specific ML framework.

\revised{
\noindent \textbf{SGML with $\bm{\mathcal{PW}}_2$.} For this final experiment we used $\mathcal{PW}_2$ instead of $\mathcal{RPW}_2$. This experiments show that $\mathcal{PW}_2$ and $\mathcal{RPW}_2$ have equivalent results. This suggests that projecting only on the canonical basis is sufficiently informative while still being less costly. }

Globally, the ablative study is in favor of the choices proposed for SGML.
The driving idea in the present article of choosing simple and scalable methods over more complex ones, leads to competitive performance while allowing full scalability.

\section{Conclusion}
\label{s:conclusion}
In this article, we proposed a metric learning method for attributed graphs, specifically to increase the performance of k-NN. We have shown experimentally that it can indeed achieve performance similar or even superior to the state of the art. 
However, a theoretical work on the properties of $\mathcal{RPW}_2$ will be useful to allow us to better understand when it does not perform  well. \oldrevised{ Appendix~\ref{ann:limitations} presents some additional elements on the limits of the work. In addition, further work may easily adapt \textbf{SGML} to perform other tasks like graph clustering or regression, with an appropriate (and probably different) ML loss.}

\section*{Acknowledgements}

The work has been was supported by the ACADEMICS Grant given by the IDEXLYON project of the Universit\'{e} de Lyon, as part of the "Programme Investissements d'Avenir" ANR-16-IDEX-0005, the ANR-19-CE48-0002 DARLING grant, and the GraphNEx CHIST-ERA ANR-21-CHR4-0009 grant.
\color{black} 

%\newpage

\bibliographystyle{unsrtnat}
\bibliography{reference}

\newpage
\appendix
\textbf{\Large Supplementary Materials for \medskip \\
``A simple way to learn metrics between attributed graphs'' \medskip \\
Yacouba Kaloga, Pierre Borgnat, Amaury Habrard} \\

\section{Appendix}

%\revised{
\subsection{Motivation of $\mathcal{RPW}_2$}
\label{annexe_rpw}
 
We give here, additional information to justify the form that $\mathcal{RPW}_2$ (Eq.~(\ref{eq:rpw})) takes. In particular, the choice of using canonical vector basis $\{u_i\}_{i=1}^p$ of $\mathbb{R}^p$ to project distributions. We will show that this choice ensures that $\mathcal{RPW}_2$ \newrevised{is a metric on discrete distribution space of $\mathbb{R}^p$, verifying symmetry, identity of indiscernibles and triangular inequality properties} just like $\mathcal{PW}_2$ from which it is derived. Then we will show that for any choice of vector family which does not span $\mathbb{R}^p$, ensuring the identity of indiscernibles may require the same amount of computation as  to calculate $\mathcal{W}_2$. \newrevised{Note that here we will not write the proof that $\mathcal{PW}_2$ is a metric since it can be derived straightforwardly from the proof that $\mathcal{RPW}_2$ is a metric. It is also easy to derive from this proof that $\mathcal{RPW}_s$ (and $\mathcal{PW}_s$)  for $s\in \mathbb{N}^{*}$ is also a metric on discrete distribution space.}

In order to facilitate the reading we recall below the definition of $\mathcal{RPW}_2$~:

\begin{equation}
\label{eq:rpw_recall}
\begin{split}
\mathcal{RPW}_2(\mu,\nu)^2 =  \frac{1}{p} \sum_{k = 1}^{p}  \sum_{i,j=1}^{n,n'} \pi_{i,j}^{ {\vect{u}_k},*} |\!|\vect{x}_i-\vect{x}_j'|\!|_2^2  
\end{split}
\end{equation}

\newrevised{Where $\mu = \sum_{i =1}^n a_i \delta_{\vect{x}_i}$ and $\nu = \sum_{i = 1}^{n'} b_i \delta_{\vect{x}_i'}$ are distributions of $\mathbb{R}^p$ with strict positive weights $(a_i)_{i=1}^n$ and $(b_i)_{i=1}^{n'}$} \newrevised{which add up to 1}; $\pi^{\vect{u}_k,*}$ are 1-d optimal transport plans of projected distributions $\mu_{\vect{u}_k} = \sum_{i = 1}^{n} a_i \delta_{\vect{x}_i(k) }$ and $\nu_{\vect{u}_k} = \sum_{i = 1}^{n'}  b_i \delta_{\vect{x}_i'(k)}$. Since $\mu_{\vect{u}_k}$ and $\nu_{\vect{u}_k}$ are projected distribution they may have non unique bins which lead to potentially several optimal transport plans, in such cases the chosen transport $\pi_{i,j}^{\vect{u}_k,*}$ is one of those which minimize Eq.~(\ref{eq:rpw_recall}).

\subsubsection{$\mathcal{RPW}_2$ \newrevised{is a metric on discrete distribution space}}
\newrevised{Let's show that $\mathcal{RPW}_2$ verifies the three canonical properties of metrics.}

\textbf{Symmetry.} The symmetry is straightforward to derive. Since the cost $|\!|\vect{x}_i-\vect{x}_j'|\!|_2^2$ and the transport plans $\pi^{\vect{u}_k,*}$ (which depend exclusively on the sorting of $\big(x_i(k)\big)_{i \in \{1,\dots,n\}}$ and $\big(x_j'(k)\big)_{j \in \{1,\dots,n'\}}$) are symmetric, $\mathcal{RPW}_2$ is thus symmetric.

\newrevised{

\textbf{Identity of indiscernibles.} We have to show that $\mathcal{RPW}_2(\mu,\nu) = 0$ if and only if $\mu = \nu$.

First,  let's assume that $\mu \neq \nu $, then:

\begin{align}
    \mathcal{RPW}_2(\mu,\nu)^2 &=  \frac{1}{p} \sum_{k = 1}^{p}  \sum_{i,j=1}^{n,n'} \pi_{i,j}^{ {\vect{u}_k},*} |\!|\vect{x}_i-\vect{x}_j'|\!|_2^2  
    \\ \text{By definition of $\mathcal{W}_2$ (Eq.~(\ref{eq:wasse}))\:\:\:\:\:\:\:\:\:\:\:\:}
    & \geq  \frac{1}{p} \sum_{k = 1}^{p}  \mathcal{W}_2(\mu,\nu)^2 \\ \text{Because $\mathcal{W}_2$ is a metric and } \mu \neq \nu \text{\:\:\:\:\:\:\:\:\:\:\:\:} &> 0
 \end{align}
Hence $\mathcal{RPW}_2(\mu,\nu) = 0 \implies \mu = \nu$.

Secondly, let's assume that $\mu = \nu$. This implies that $n = n'$, $a_i = b_i $ and $\mu_{\vect{u}_k} = \nu_{\vect{u}_k}$ for $ i \in \{1,\dots,n\}$ and $k \in \{1,\dots,p\}$. Because of these equalities, each optimal transport $\pi^{\vect{u}_k,*}$ can be associated to a unique permutation. By introducing the notation $\sigma_k$ such that $\forall i,j \in \{1,\dots,n\}, \sigma_k(j) = i $ iff $\pi_{i,j}^{\vect{u}_k,*}  = a_j$, we can write:

\begin{equation}
% \label{eq:rpw_rewrite}
\begin{split}
\mathcal{RPW}_2(\mu,\nu)^2 =  \frac{1}{p} \sum_{k = 1}^{p}  \sum_{j=1}^{n}  a_j|\!|\vect{x}_{\sigma_k(j)}-\vect{x}_j'|\!|_2^2  
\end{split}
\end{equation}

In addition, since $\mu = \nu$, they have the same bins, therefore there is a permutation $\tau$ such that:

\begin{equation}
% \label{eq:rpw_rewrite_}
\begin{split}
\vect{x}_{\tau(j)} = \vect{x}_j'
\end{split}
\end{equation}

Thus: 

\begin{equation}
% \label{eq:rpw_rewrite}
\begin{split}
 \forall k \in \{1,\dots,p\}, \;\; \sum_{j=1}^{n}  a_j |\!|\vect{x}_{\tau(j)}(k) -\vect{x}_j'(k)|\!|_2^2  = 0
\end{split}
\end{equation}

Since, $\forall k \in \{1,\dots,p\}$: 

\begin{equation}
% \label{eq:rpw_recal}
\begin{split}
 0 = \mathcal{W}_2(\mu_{\vect{u}_k},\nu_{\vect{u}_k}) = \sum_{j=1}^{n} a_i |\!|\vect{x}_{\tau(j)}(k) -\vect{x}_j'(k)|\!|_2^2 
\end{split}
\end{equation}

It is clear that $\tau$ is a permutation associated to an optimal transport between $\nu_{\vect{u}_k}$ and $\mu_{\vect{u}_k}$. Therefore by definition of $\sigma_k$~ (i.e. a permutation associated with the optimal transport between $\nu_{\vect{u}_k}$ and $\mu_{\vect{u}_k}$ which minimizes Eq.~(\ref{eq:rpw_recall})):

\begin{equation}
\label{eq:rpw_rewrite__}
\begin{split}
\mathcal{RPW}_2(\mu,\nu)^2  =  \frac{1}{p} \sum_{k = 1}^{p}  \sum_{j=1}^{n}  a_j |\!|\vect{x}_{\sigma_k(j)}-\vect{x}_j'|\!|_2^2  \leq  \frac{1}{ p} \sum_{k = 1}^{p}   \sum_{j=1}^{n} a_j |\!|\vect{x}_{\tau(j)}-\vect{x}_j'|\!|_2^2  = 0
\end{split}
\end{equation}

Hence $\mu = \nu \implies \mathcal{RPW}_2 = 0 $. This concludes the proof.

\textbf{Triangular inequality.} Let's consider the distributions  
$\mu = \sum_{i = 1}^n a_i \delta_{\vect{x}_i}$,  
$\nu = \sum_{i = 1}^{n'} b_i \delta_{\vect{x}_i'}$, 
and a third distribution $\zeta = \sum_{i = 1}^w c_i\delta_{\vect{z}_i} \in \mathcal{P}(\mathbb{R}^p)$ 
with strictly positive coefficient $(c_i)_{i=1}^w$. We have to show that $\mathcal{RPW}_2(\mu,\nu) \leq  \mathcal{RPW}_2(\mu,\zeta) + \mathcal{RPW}_2(\zeta,\nu)$. 

In order to ease the reading of the proof, we will make an abuse of notation. For any k $\in \{1,\dots,p\}$, we denote $\pi_{i,l}^{\vect{u}_k,*}$ (resp. $\pi_{l,j}^{\vect{u}_k,*}$) the coefficients of optimal transport plan between 
$\mu_{\vect{u}_k}$ and $\zeta_{\vect{u}_k}$ (resp. $\zeta_{\vect{u}_k}$ and $\nu_{\vect{u}_k}$). 

Since $\mu_{\vect{u}_k}$, $\nu_{\vect{u}_k}$ and $\zeta_{\vect{u}_k}$ are 1-d dimensional distributions, the composition of the optimal transport plans between $\mu_{\vect{u}_k}$ and $\zeta_{\vect{u}_k}$ and the transport plan between $\zeta_{\vect{u}_k}$ and $\nu_{\vect{u}_k}$ is an optimal transport plan between $\mu_{\vect{u}_k}$ and $\nu_{\vect{u}_k}$. This one is expressed as  $ \pi_{i,j}^{\vect{u}_k,*} = \sum_{l=1}^{w } \pi_{i,l}^{\vect{u}_k,*}  \pi_{l,j}^{\vect{u}_k,*} /c_l $.

Therefore by definition of $\mathcal{RPW}_2$, we can write : 

\begin{align}
\mathcal{RPW}_2(\mu,\nu) &\leq \Bigg( \frac{1}{p}\sum_{k=1}^p\sum_{i,j=1}^{n,n'} \pi_{i,j}^{\vect{u}_k,*}  |\!|\vect{x}_{i}-\vect{x}_j'|\!|_2^2 \Bigg)^{\frac{1}{2}}  \nonumber \\
&\leq \Bigg( \frac{1}{p}\sum_{k=1}^p\sum_{i,j=1}^{n,n'} \sum_{l=1}^{w } \frac{\pi_{i,l}^{\vect{u}_k,*}  \pi_{l,j}^{\vect{u}_k,*} }{c_l} |\!|\vect{x}_{i}-\vect{x}_j'|\!|_2^2 \Bigg)^{\frac{1}{2}}  \nonumber \\
&\leq \Bigg(\frac{1}{p}\sum_{k=1}^p\sum_{i,j,l=1}^{n,n',w}  \frac{\pi_{i,l}^{\vect{u}_k,*}  \pi_{l,j}^{\vect{u}_k,*} }{c_l} |\!|(\vect{x}_{i}-\vect{z}_l)+(\vect{z}_l-\vect{x}_j')|\!|_2^2 \Bigg)^{\frac{1}{2}}  \nonumber\\
\end{align}

Using the Minkowski inequality, we have:

\begin{align}
\mathcal{RPW}_2(\mu,\nu) & \leq\Bigg(\frac{1}{p}\sum_{k=1}^p \sum_{i,j,l=1}^{n,n',w}     \frac{\pi_{i,l}^{\vect{u}_k,*}  \pi_{l,j}^{\vect{u}_k,*} }{c_l} |\!|\vect{x}_{i}-\vect{z}_l |\!|_2^2 \Bigg)^{\frac{1}{2}}+\Bigg( \frac{1}{p}\sum_{k=1}^p\sum_{i,j,l=1}^{n,n',w} \frac{\pi_{i,l}^{\vect{u}_k,*}  \pi_{l,j}^{\vect{u}_k,*} }{c_l} |\!| \vect{z}_l  -  \vect{x}_j'|\!|_2^2\Bigg)^{\frac{1}{2}}\nonumber\\
& \leq\Bigg( \frac{1}{p}\sum_{k=1}^p\sum_{i,l=1}^{n,w}  \pi_{i,l}^{\vect{u}_k,*}     \frac{\sum_{j=1}^{n'}\pi_{l,j}^{\vect{u}_k,*} }{c_l} |\!|\vect{x}_{i}-\vect{z}_l |\!|_2^2 \Bigg)^{\frac{1}{2}}+\Bigg( \frac{1}{p}\sum_{k=1}^p\sum_{j,l=1}^{n',w} \frac{\sum_{i=1}^{n}\pi_{i,l}^{\vect{u}_k,*}   }{c_l} \pi_{l,j}^{\vect{u}_k,*}|\!| \vect{z}_l  -  \vect{x}_j'|\!|_2^2\Bigg)^{\frac{1}{2}}\nonumber\\
\end{align}
Since by definition $c_l = \sum_{j=1}^{n'}\pi_{l,j}^{\vect{u}_k,*}  = \sum_{i=1}^{n}\pi_{i,l}^{\vect{u}_k,*}$, this leads to:

\begin{align}
\mathcal{RPW}_2(\mu,\nu) & \leq\Bigg( \frac{1}{p}\sum_{k=1}^p \sum_{i,l=1}^{n,w}   \pi_{i,l}^{\vect{u}_k,*} |\!|\vect{x}_{i}-\vect{z}_l |\!|_2^2 \Bigg)^{\frac{1}{2}}+\Bigg(\frac{1}{p}\sum_{k=1}^p \sum_{j,l=1}^{n',w}  \pi_{l,j}^{\vect{u}_k,*}  |\!| \vect{z}_l  -  \vect{x}_j'|\!|_2^2\Bigg)^{\frac{1}{2}}\nonumber\\
&\leq \mathcal{RPW}_2(\mu,\zeta) + \mathcal{RPW}_2(\zeta,\nu)  \nonumber
% \end{split}
\end{align}

This concludes the proof. Therefore $\mathcal{RPW}_2$ is a metric for the discrete distributions on $\mathbb{R}^p$.

}

%\end{document}

\subsubsection{Projection on canonical basis vector}

Projecting onto the family of canonical vector basis is a choice that stems from a spanning constraint, and also a choice of simplicity. As we stated, for a given $k$, there may be several optimal transport plans between $\mu_{\vect{u}_k}$ and $\nu_{\vect{u}_k}$. This happens when, for a given $k$, there are bins where~:

\begin{equation}
\label{eq:condition}
\vect{x}_i(k) = \vect{x}_j(k) \text{\:\:\:\:\:and\:\:\:\:\:} \vect{x}_i \neq \vect{x}_j
\end{equation}

For continuous data (such as those that are the outputs of a GCN) this condition is particularly unlikely to happen. However, if we were projecting onto a non spanning vector family, that we call $\big\{v_u\big\}$, every bins lying in an orthogonal space of $\textbf{span}\big\{v_u\big\}$ would be projected on 0. This would lead on several couple of bins where the above condition (\ref{eq:condition}) would be verified. In an extreme case where all bins of $\mu$ and $\nu$ would be in the orthogonal space, any transport plan would be optimal between projected distribution. Therefore, finding the OT plan which minimizes (\ref{eq:rpw_recall}) would be equivalent to finding the optimal transport plan of Wasserstein distance. This is the reason why it is mandatory to find a family of vectors which spans $R^p$. 

A natural choice is then to use the canonical basis from which it is easy to derive the identity of indiscernibles property (proven above) and from which the projection is costless on a numerical point of view. Anyway, in a different context from this work, where the distribution bins are not only continuous but also fit categorical data, some other spanning family may be more suitable. We hope that future works will take this idea to use specific family of vector to build specific distance (independently of the question of approximation of an existing OT distance) a step further.

{\color{black}
\subsection{Motivation and Interpretation of NCCML}
\label{ann:loss}

We detail here some of the insights that led us to propose NCCML for ML.

Since we want to maintain a low complexity to train our model, a batch training is desirable. As a consequence and as said in section~\ref{ss:loss}, the Large Margin Nearest Neighbor (LMNN) \citep{weinberger09} loss was not appropriate because it works very locally and is not optimal with batch training. Indeed, LMNN tries to attract and repel points with elements of the datasets which are neighbours, according to their labels. On a batch training, this could lead to some \textit{overfitting} where we try to attract points which should not be close even if they share the same label. This is even true the smaller the batch size we use. 

An alternative is to use Neighborhood Component Analysis (NCA) \citep{goldberger05} which provided a slightly better but limited performance. In reality, NCA is also a very locally method. Indeed it considers the probability $p(\mathcal{G}_i,\mathcal{G}_j)$ for two elements to have the same labels:

\begin{equation}
\label{eq:nca2}
p_{\mat{\Theta}}(\mathcal{G}_i,\mathcal{G}_j) = \frac{\exp \Big({ - d_{\mat{\Theta}}^{\mathcal{RPW}_2} {(\mathcal{G}_j,\mathcal{G}_i)^2}}\Big)}{ \sum_{k,k'} \exp \Big({ - d_{\mat{\Theta}}^{\mathcal{RPW}_2} {(\mathcal{G}_k,\mathcal{G}_{k'})^2}}\Big)}
\end{equation}

 Given this form of probability, it tries to maximize them for all elements which have effectively the same labels:

\begin{equation}
\label{eq:nca}
\max_{\mat{\Theta}}  \sum_{\mathcal{G}_i \in \mathbb{G} }  \sum_{\substack{ \mathcal{G}_j \in \mathbb{G} \\ \mathcal{E}(\mathcal{G}_i) = \mathcal{E}(\mathcal{G}_j)  }}  p_{\mat{\Theta}}( \mathcal{G}_i , \mathcal{G}_j) 
\end{equation}

However, as one can see from Eq.~(\ref{eq:nca}), the probability of having the same labels is a softmax, so distant elements do not contribute a lot to these probability. It contains mostly local information. We believe that one could obtain better results by considering a more global criterion. Moreover, using a batch would be now advantageous since it will help the model to build good metric, even for k-NN (which requires a local fine metric) since the batch training will act as a regularization and will help to generalize.

An inspiration for that comes from NCMML~\citep{mensink12} which proposes a loss function specifically built to increase performance of nearest mean classifier. This model also relies on a probabilistic model where the probability to belong on a class is given by a softmax which considers the distance to the mean of different classes. Obviously NCMML is not well suited for our tasks using kNN. Plus, it would require an additional layer of computation for computing barycenter with OT.

We took a compromise between NCA and the NCMML loss. The probability to be part of a class is given by a softmax which depends on the relative distance to different same label element (Eq.~(\ref{eq:ncmml2})). It has the advantage that the loss on a batch will be representative of the loss over the whole dataset, because the relative distance to different labels should remain the same also on subsamples of the dataset. Moreover it benefits from the batch training which acts as a regularizer. That finally leads to a better metric learned compared to NCA for k-NN as proven on our ablative study (Table.~\ref{data_results2}). Anyway, in a regular setting where we could use all datasets to build and train theses losses, NCCML would certainly shows worse results than LMNN and NCA. 

The specific settings that is studied here, due to the requirement of  scalability, forces to propose a loss different from the literature, that indeed brongs some improvement when compared to NCA.

}

\subsection{Implementation details}
\label{ann:imp}
\textbf{Sequential implementation.} A priori, it is necessary to compute all the transport costs between two distributions so as to calculate the optimal transport and this operation has a quadratic complexity. For most OT distance such as $\mathcal{W}_2$, since the complexity is dominated by the computation of the optimal transport plan, this was of no consequence. However for $\mathcal{RPW}_2$ (as well as for $\mathcal{SW}_2$) it becomes a critical aspect. Hopefully, there is no need to compute all the costs to find the optimal transport and the transport cost has no more than $n+m$ (given that the distributions have sizes $n$ and $m$) non zero coefficients. This is why their complexity remains quasi-linear in $O(n\log n)$. The algorithm of the implementation referred to as "sequential implementation" in the core text can be found on Algorithm~\ref{algo:rpswseq}. The experiment on Section~\ref{sec:rtc} assessed the quasi-linear complexity of this algorithm.

\textbf{Quadratic implementation.} In this second implementation, we compute all possible transport costs using a library of matrix multiplication, and then we multiply these costs by the optimal transport matrix. These operations allow us to benefit from the advantages of vectorization and to gain time compared to the sequential implementation, when $n$ is not too large. This result is assessed experimentally in Section~\ref{sec:rtc}.

Both implementation can be found with this supplementary material.

\noindent \textit{\textbf{Note:} In the reported experiments, we have seen that for $n < 1000$, it's better to use the quadratic implementation. Anyway this result strongly depends on the hardware used, and also on the dimension of the distribution support $p$. 
The scaling behavior of the two implementations is an interesting characteristic, showing than the proposed method can be implemented in a quasi-linear way. The second comment is also that the method can be made rapid enough (and very competitive) with optimizations.}

\begin{algorithm}[h]
\caption{ $\mathcal{RPW}_2$ - \text{Sequential}} 
\label{algo:rpswseq}
\begin{algorithmic} 
\ENSURE Build the distance between two discrete distributions  $\mu$ and $\nu$ in $\mathcal{P}(\mathbb{R}^{p})$.
\REQUIRE $\mu = \sum_{i = 1}^n a_i \delta_{\vect{x}_i}$ and $\nu = \sum_{j = 1}^m b_i \delta_{\vect{y}_j}$.
\STATE Set $c = 0$.
\FOR{each epoch $k \in \{1,\dots,p\}$}
\STATE Get $\sigma_\mu^k$, $\sigma_\nu^k$ sort permutation of supports vectors $k$-th components.
\STATE i.e  $\vect{x}_{\sigma_\mu^k(0)}(k) \leq \dots \leq \vect{x}_{\sigma_\mu^k(n-1)}(k)  $ and $\vect{y}_{\sigma_\nu^k(0)}(k) \leq \dots \leq \vect{y}_{\sigma_\nu^k(m-1)}(k) $.
\STATE Set $T =$ \textit{true}. Set $i,j = 0, 0$.
\STATE Set $w_\mu,w_\nu = a_{\sigma_\mu^k(0)}, b_{\sigma_\nu^k(0)}$.

\WHILE{$T$ == \textit{True}}
\IF{$w_\mu < w_\nu$}
\STATE $c = c + w_\mu*|| \vect{x}_{\sigma_\mu^k(i)} - \vect{y}_{\sigma_\nu^k(j)} ||_2^2$
\STATE $i= i+ 1$
\IF{$i == n$}
\STATE $T=$ \textit{false}
\ENDIF
\STATE$w_\nu = w_\nu - w_\mu$
\STATE $w_{\mu} = a_{\sigma_\mu^k(i)}$
\ELSE
\STATE $c = c + w_\nu*|| \vect{x}_{\sigma_\mu^k(i)} - \vect{y}_{\sigma_\nu^k(j)} ||_2^2$
\STATE $j= j+ 1$
\IF{$j == m$}
\STATE $T=$ \textit{false}
\ENDIF
\STATE$w_\mu = w_\mu - w_\nu$
\STATE $w_{\nu} = b_{\sigma_\nu^k(j)}$
\ENDIF
\ENDWHILE
\ENDFOR
\RETURN $\sqrt{\frac{c}{q}}$
\end{algorithmic}
\end{algorithm}

\subsection{Datasets}
\label{app:datasets}
The characteristics of the datasets used are summarized in Table \ref{data_sets}.

\begin{table}[t]
\centering
\caption{ \textbf{Graph datasets used in our experiments.}  \#Graphs: number of graphs. \#Nodes: average number of nodes. cont.: attributes have continuous values; lab.: attributes are labels. deg.: the featurattributes are degrees of nodes. $q$ is the feature dimension.
\label{data_sets}}
 \setlength{\tabcolsep}{2.25pt}
 \fontsize{9}{10}\selectfont
\begin{tabular}{cccccccccc}
\toprule
Datasets & BZR    & COX2 & {PROTEINS}  & {ENZYMES}    & MUTAG    & NCI1    & IMDB-B& IMDB-M& CUNEIFORM  \\ 
\midrule
\#Graphs  &   405     &   467  & {1113}   &600   &   188       &  4110    & 1000 & 1500& 267 \\  
\#Nodes  &   35.75     &   41.22  & {39.06}    & 32.63   &   17.93      &  29.97    & 19.77  & 13& 21.27  \\
Node attributes  &   cont.     &   cont.  & cont. / lab. & cont. / lab.  &   deg.    &  lab.    & deg.& deg. & cont. / lab.  \\
$q$  &   3     &   3  & 1 / 3   &  18 /3   &   4      &  38    &  135 & 88& 3 / 3 \\
\bottomrule
\end{tabular}
\end{table}

\begin{table}[t]
\centering
\caption{ \textbf{Typical runtimes in our experiments.} The running time of WWL ($r = 2$) and $\mathcal{FGW}$ ($\alpha = 0.5$ except for IMDB datasets where it is set to $1$) to calculate distances are also provided. 
\label{data_sets_runtimes}}
 \setlength{\tabcolsep}{2.25pt}
 \fontsize{9}{10}\selectfont
\begin{tabular}{cccccccccc}
\toprule
Datasets & BZR    & COX2 & PROTEINS [lab.] & ENZYMES [lab.]   & MUTAG    & NCI1    & IMDB-B/M)& IMDB-M \\ 
\midrule
Training time (s)  &   35    &    40  & 240 &  220   &   15       &  480    & 80/120 & 120\\  
Distances comp. (s)   &   5     &   7   & 40 &  40  &   1     &  480    & 10/55  & 55 \\
\bottomrule
\oldrevised{Dist. comp. (s) -  WWL}   &   16     &   25   & 200 &  30  &   2     &  1500    & 80/140  & 140 \\
\oldrevised{Dist. comp. (s)  $\mathcal{FGW}$} &   240      &   270   & 1h&  540  &   30     &  6h30min    &  1000/1400  & 1400 \\
\bottomrule
\end{tabular}
\end{table}

\subsection{\textbf{SGML} - Datasets runtimes}
\label{annexe_runtimes}
The following Table~\ref{data_sets_runtimes} provides the typical runtimes for both training part and distance computation phases for the different datasets considered in this paper. We used the proposed quadratic implementation for all datasets. A \texttt{tensorflow} implementation is used during the training phase (to leverage the build-in functions for optimization and training) while the \texttt{numpy} implementation is used during the final distance computation. All running time experiments were conducted with a computer equipped with an Intel CORE i9900ks processor (62 GB of RAM) and GeForce RTX 3090 (24 GB of RAM). 

The parameters are the same as in the experiments described in the paper. We fixed the depth of our GCN to $r=4$. 

As we can see despite lower theoretical complexity, the training time is bigger than distance computation. This is because the  \texttt{numpy} implementation is much more efficient (especially in computing the sort operation) and these datasets are not large enough (in terms of the number of graphs) for \texttt{tensorflow} implementation  catches up to  \texttt{numpy} implementation. One clearly sees that the bigger the dataset (e.g., the NCI1 dataset), the lower the \texttt{numpy} implementation saves time.

\begin{figure}[h]
    \centering
    \includegraphics[width=13cm,height=29cm,keepaspectratio]{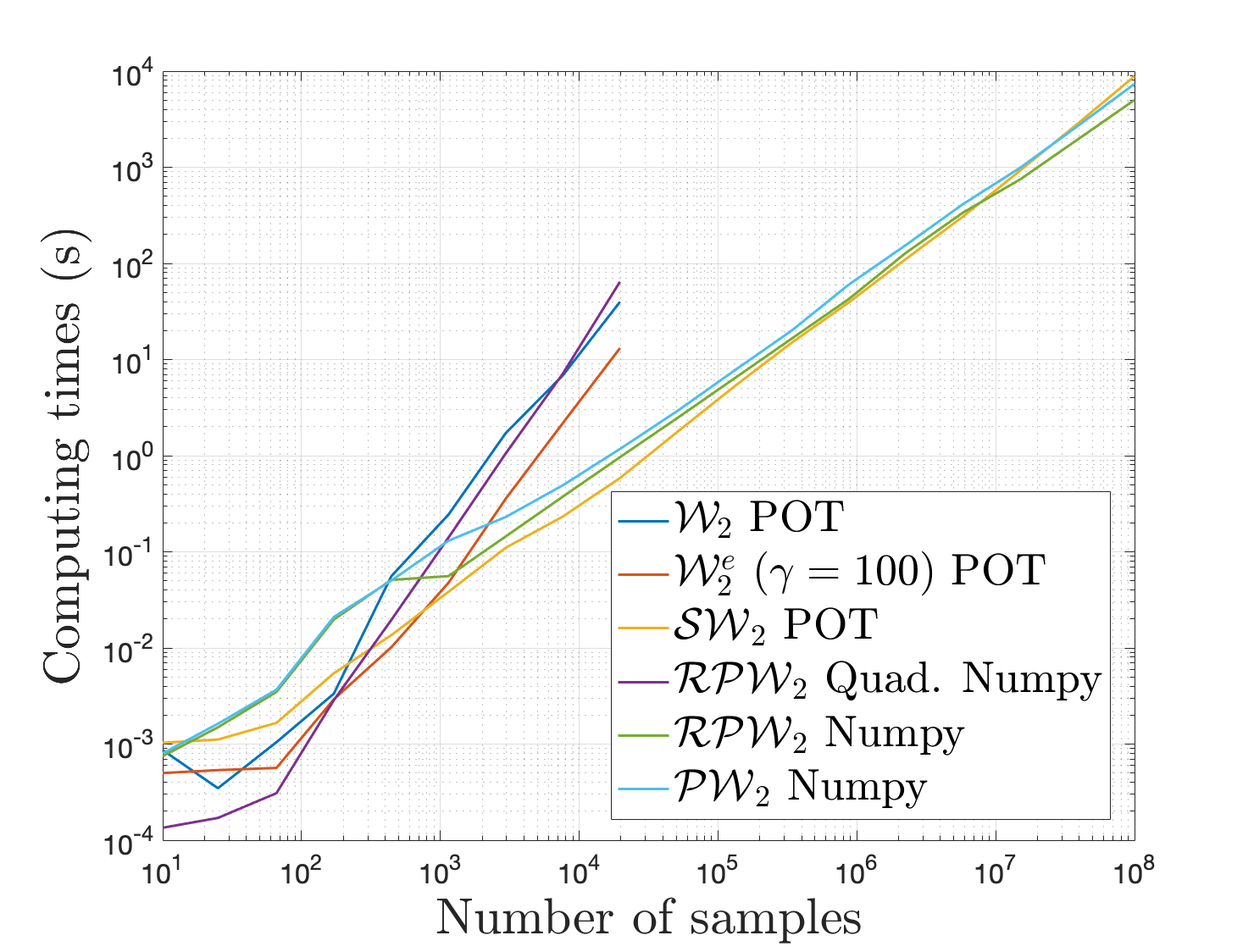}
    \caption{\color{black} \textbf{Run time comparisons 2.}}
    \label{fig:rnt2to}
\end{figure}

{\color{black}
\subsection{$\mathcal{RPW}_2$ runtimes according to graph size}
\label{annexe_runtimes2}
In section \ref{sec:rtc}, we have not been able to extend the comparison of computation times between $\mathcal{RPW}_2$ and $\mathcal{SW}_2$ up to $10^8$ size distributions in the same experimental  conditions. The reason is that, beyond approximately 6 million points, we encountered a memory issue with $\mathcal{SW}_2$ on our Intel CORE i9900ks processor $\times$ 62 GB of RAM computer. It appears to be an implementation issue from POT toolboxes. Anyway, we have redone all the experiment with a computer with more RAM but a less powerful processor, an Intel Xeon Gold 5218 $\times$ 2 To of RAM. This amount of RAM is obviously overkill but it allows us to avoid any issue on the $\mathcal{SW}_2$ implementation. The results can be found in Figure~\ref{fig:rnt2to}. It confirms the calculated complexity on Section \ref{sec:ca}, asymptotically  $\mathcal{RPW}_2$ scale better than $\mathcal{SW}_2$ since in our settings $p (= 5) < M (= 50)$.
}

\subsection{Additional details}
\label{ann:adddetails}
\noindent \textbf{ENZYMES.} (discrete) The learning rate $0.999 \:10^{-2}$ was too heavy for NCA loss on ENZYMES, so we used $0.999\:10^{-3}$ for this dataset. Accordingly we set the number of epochs to 20. However, we let the possibility to early stop at 10 epochs, meaning that the epochs number $E$ becomes an hyper-parameter. $E=\{10,20\}$. %in the same way as $r$.  }

\textbf{PROTEINS.} The above remark applies to PROTEINS (with continuous attributes). The learning rate was set to $0.999 \:10^{-4}$ and the epochs number $E$ becomes an hyper-parameter $E=\{10,20\}.$ 
\oldrevised{

\textbf{CUNEIFORM.} Since it has 30 different labels, the batch size has been set to 64. 

\noindent \textbf{ENZYMES.} (continuous) It was trained in the same way as ENZYMES (discrete).

\textbf{Extended vector attributes.} We used a concatenation of continuous attributes and one-hot encoding of discrete attributes to build an extended vectors attributes. Since our method is a ML method it is pertinent to give all information we have and let the method to select the most relevant information. In case of PROTEINS, this choice was motivated because its node features are  scalars which is not suitable for the adaptation procedure while in case of ENZYMES (continuous) and Cuneiform using only continuous attributes lead to poor results. This choice help to have more flexibility for SGCN to build the metric while avoiding to use more powerful but also more costly GCN.}

{\color{black}
\subsection{FGW with k-NN}
\label{sec:abla_fgw}
In the ablative study, we evaluated WWL with a k-NN to justify the design choice. Here, as a complement, we reproduced this experiment with  $\mathcal{FGW}$.  $\mathcal{FGW}$ has a parameter denoted $\alpha\in [0,1]$ which sets the trade-off between the structure and the characteristics of the nodes in the distance computation. We performed a small grid search over this parameter $\alpha = [0.25,0.5,0.75]$. Except for IMDB datsaets where $\alpha = 1$ as in original paper. The results can be found in Table~\ref{data_results3}. One can see that the results are mitigated, FGW performs very well on some datasets and much less well on others. Moreover one could probably get even better results by doing a much larger hyperparameters tuning, as in the $\mathcal{FGW}$ original paper. Still, the present comparison is fair since, first, the grid search on the proposed method was also relatively small. Second, these results must be analyzed keeping in mind the significant difference in calculation time between the two methods (see Table~\ref{data_sets_runtimes}). This illustrates also that doing a fine hyperparameter tuning with such expensive methods is not often feasible on very large data sets.

}
\begin{table}[t]
  \caption{\oldrevised{\textbf{Ablative experiment with $\mathcal{FGW}$.} Acc. is the accuracy. $\Delta$ is the difference in accuracy between the model of the column and the proposed one SGML whose results are on Table.~\ref{data_results}. Red negative (resp. Green positive) number means that our model perform better (resp. worse). {\xmark} $\:$symbol means that we had infinite distance values with the default settings of FGW solver.}}
\label{data_results3}

 \setlength{\tabcolsep}{2.25pt}
 \fontsize{9}{10}\selectfont

  \centering
  \begin{tabular}{ll l}
    \toprule
  Dataset     & \multicolumn{2}{c}{\textbf{ $\mathcal{FGW}$}}  \\
  \cmidrule(r){2-3} 

  Method & \textbf{Acc.}   & $\bm{\Delta}$   \\
  \midrule
  \textbf{BZR}    & 81.70    &  \color{red} - 3.91   \\ 
\textbf{COX2}   &  78.51   &   \color{red} - 1.28      \\
\textbf{MUTAG}  &  83.16 & \color{red}  - 6.84      \\ 
\textbf{NCI1} &    \xmark  & \xmark    \\ 
\textbf{PROTEINS}  &   \xmark    &   \xmark   \\ 
\textbf{IMDB-B}  &   80.80      &  \color{green}   11.9 \\%
\textbf{IMDB-M}  &  \xmark    &   \xmark  \\ 
\textbf{ENZYMES} &  70.83 &  \color{green}  19.33  \\ 
    \bottomrule 
  \end{tabular}
\end{table}

{\color{black}
\subsection{Limitations of this study}
\label{ann:limitations}

We discuss some of the limitations of the model and give some suggestions for improvements.

\noindent \textbf{GCNs}. To generate the distributions associated with the graphs, the model relies on a Graph Convolutionnal Neural network (GCN).

Because of this we can expect some sub-optimal behavior of  the model in terms of expressiveness. Indeed, while they are very efficient to characterize graphs locally, GCNs tend to lose efficiency when their depths increase. Although variations on their architectures have been proposed to solve this issue~\cite{luan2019break}, it appears that most neural networks show similar results~\cite{wu19} and this defect seems to be intrinsic of their low-pass message passing scheme~\cite{loukas_19}. Therefore, new ways to efficiently characterize graphs at small and large scales could allow learning a better metric. In this regard, transformers are promising methods~\cite{yun19,vijay20}. Their ability to characterize context at different scales has already been successfully exploited in natural language processing tasks. Currently many attempts have been made in recent years to adapt them to graphs. However, these are difficult networks to train and their integration in SGML would not result in a simple and scalable metric learning model.

\noindent \textbf{Performance}. The model allows us to obtain an improvement in classification with k-NN as compared to the current methods. It is then more suitable for dealing with real datasets where new input are available after (or coming as graph streams) as the method do not need to be fully re-trained. However, performance with the k-NN remain inferior to those reached with a SVM. Thus in a critical real application (medical for example), where performance is of utmost importance, it is preferable to use the SVM. Additional work would be therefore necessary to gain more performance with the k-NN. This gain in performance could be acquired by introducing a different model of GCN so to generate the features, as mentioned above. But it could also done by making the model more complex. For example instead of considering uniform distributions from GCN features, we could introduce an attention mechanism that could modulate theirs weights on the distributions. This could give more flexibility to the model to build the metric, but at the cost of a more expensive training.  

\noindent \textbf{Theoretical}. The work on the distance that we introduced here, $\mathcal{RPW}_2$, which is scalable and has a good behavior in our model, is currently methodological and driven by insight. As of today,  we have not proven that it satisfies the triangular inequality so that it is not guaranteed that it is a true metric or not. This aspects remains to be clarified.

\noindent \textbf{Opening to other tasks}. Our work has been limited here to the k-NN for supervised classication. But other relevant classifiers with interesting properties where NCCML is not efficient enough could be consider, eg. the Nearest Class Mean~\cite{mensink13}. Other tasks can also be considered such as clustering (k-means, ...) and regression (k-NN regression, ...). We believe that the present work is a first step on the goal of lowering the cost of many other tasks on graphs.
}

\end{document}